\RequirePackage{fix-cm}

\documentclass[smallcondensed]{svjour3}
\smartqed 

\usepackage{times}

\usepackage{soul}
\usepackage{url}
\usepackage[utf8]{inputenc}
\usepackage{caption}
\usepackage{dsfont}
\usepackage{subcaption}
\usepackage{graphicx}
\usepackage{amsmath}
\usepackage{booktabs}
\usepackage{multirow}
\usepackage[table]{xcolor}
\usepackage[margin=1.3in]{geometry}
\usepackage{cancel}

\usepackage{apacite}

\begin{document}

\title{Addressing Fairness, Bias and Class Imbalance in Machine Learning: the FBI-loss.}
%\subtitle{subtitle}

\author{Elisa Ferrari         \and
        Davide Bacciu %etc.
}

\providecommand{\keywords}[1]{\textbf{\textit{Index terms---}} #1}

\institute{E. Ferrari \textit{corresponding author} \at
              Scuola Normale Superiore, Pisa, Italy \\
              \email{elisa.ferrari@sns.it}           %  \\
           \and
           D. Bacciu \at
              University of Pisa, Department of Computer Science, Pisa, Italy
}

\date{Received: date / Accepted: date}

\maketitle

\begin{abstract}
Resilience to class imbalance and confounding biases, together with the assurance of fairness guarantees are highly desirable properties of autonomous decision-making systems with real-life impact.
Many different targeted solutions have been proposed to address separately these three problems, however a unifying perspective seems to be missing. With this work, we provide a general formalization, showing that they are different expressions of unbalance.
Following this intuition, we formulate a unified loss correction to address issues related to Fairness, Biases and Imbalances (FBI-loss).
The correction capabilities of the proposed approach are assessed on three real-world benchmarks, each associated to one of the issues under consideration, and on a family of synthetic data in order to better investigate the effectiveness of our loss on tasks with different complexities.
The empirical results highlight that the flexible formulation of the FBI-loss leads also to competitive performances with respect to literature solutions specialised for the single problems.
%The empirical results highlight how the FBI-loss can achieve performances that are competitive with respect to other solutions from literature targeted for each specific problem.
\end{abstract}

\section{Introduction}
The appeal of Machine Learning (ML) lies much in its ability, given sample data, to automatically discover complex patterns that work well also for out-of-sample predictions \shortcite{mullainathan2017machine}, requiring minimal data domain expertise.
However, the {\it learning} part in ML typically amounts to training a model to minimize a cost function on a known and limited sample. Thus, we could state that the outcome of the learning process is driven by the selection of the training dataset, by the definition of the cost function, and by the choice of the algorithm used to optimise the latter. While in the majority of applications these choices are quite straightforward, there are three widely agreed conditions in which conscious and non-trivial interventions are necessary: Class Imbalance (CI) \shortcite{ali2015classification}, Confounding Bias (CB) \shortcite{adeli2019bias} and Unfair Classification (UC) \shortcite{mehrabi2019survey}.\\
CI occurs when one class in the training set is overrepresented with respect to the others. This imbalance is critical because errors on the majority class can overwhelm and mask those on the minority class, leading to models that predict almost uniquely the majority category, at the expense of the others. In the most severe cases, the process of errors' minimization could favor a model that exclusively predicts the overrepresented class. 
CI is widely documented in biomedical applications as well as in fault detection, where diseases, pathological conditions and faults often present very skewed class distributions \shortcite{ali2015classification}. It can also be easily spotted in machine vision \shortcite{lin2017focal}, e.g. when pixels are classified as a rarely occurring event (i.e. presence of an object) or background (the majority class). \\
CB occurs when the distribution of the target classes in the training set is biased with respect to a non-target variable, called confounder, which instead is not related to the expected real distribution of the classes. This bias can mislead the learning process, leading to a model that bases its predictions on the associations between the input data and the confounder.
The misleading effect of unwanted biases is a well known problem in causal inference \shortcite{bareinboim2012controlling,kuroki2014measurement}, but it has only recently gained attention in the ML context.
In fact, much of earlier effort in ML focused on achieving generalization in predictive performance, which has been historically treated as distinct from causal investigations \shortcite{kreif2019machine}. However, in many ML applications, the objective is causal in nature, even if not specifically framed as such \shortcite{castro2020causality}. For instance, the paper by Caruana {\it et al.} \shortcite{caruana2015intelligible} describes a model to predict the probability of death for pneumonia, trained on medical records of patients who have previously had pneumonia. Counter-intuitively, the model found that asthma lowers the risk of death, while it is known to be a severe condition in subjects with pneumonia.
This misleading effect occurred because the patients with asthma in the training set received more care by the hospital system. In this example, the association learned from the dataset was correct, but clearly the aim of this application was to find only causal relations useful to prioritize care for patients with pneumonia.\\
If causal reasoning is only recently spreading in the ML context, the necessity to understand the reasons behind predictions has been discussed in the new field of Explainable Artificial Intelligence (XAI). In this context, Ribeiro {\it et al.} \shortcite{ribeiro2016should} developed a prediction explanation technique and showed how it can be used to check whether a model's predictions are due to spurious correlations in the training set. For example, they trained an image classifier to distinguish between wolves and huskies and showed that the predictions of the model were based on the presence of snow in the background of the picture, which was more likely in the wolves images. In another XAI-related paper, Ross {\it et al.} \shortcite{ross2017right} introduced the concept of a model {\it right for the right reasons} and developed a method for producing and explaining multiple classifiers that are accurate for qualitatively different reasons, leaving the responsibility to choose the right one to a domain expert.\\
A special case of CB is UC, that occurs when the bias involves a sensitive feature and thus the resulting model is unfair towards a protected group. Despite being well known that UC derives from the bias problem (a recent review lists 23 different types of biases causing unfairness \shortcite{mehrabi2019survey}), given the increasing number of ML applications taking life-changing decisions \shortcite{ajunwa2016hiring}, fairness issues attracted more attention than CB, forming a distinct branch of studies.
The societal impact of UC pushed the development of solutions based on the imposition of fairness properties to the classifier, while the recent works to address CB usually focus on the causes of the problem, trying to avoid learning from the confounder variable.\\
Summarizing, CI, CB and UC are conditions in which, without any active corrective intervention, the minimization of the prediction error on the training set may produce a model that does not satisfy the purposes for which it was developed: either because it is not informative with respect to a class (the CI case), or because it has learned a wrong pattern and thus it is not generalizable (the CB case) or finally because the predictions are discriminatory towards a protected group (the UC case).\\
Usually, these three conditions are studied separately and the corrections developed for them are based on different strategies. Instead, in this paper we show that they are different expressions of unbalance: Class Imbalance occurs when the target classes are unbalanced with respect to each other, while when they are unbalanced with respect to another variable we talk about Confounding Bias. Finally, if this bias involves protected attributes we refer to it as Unfair Classifications.\\
Given that unbalance is the common cause of all the three conditions, we introduce a new unified corrective loss function to address Fairness, Bias and Imbalance problems (FBI-loss).
We experimentally test the performances of our loss on six benchmarks comprising synthetic and real-world data, affected by each issue under consideration in this work, confronting it with other solutions specifically tailored for each problem.\\
In synthesis, the contributions of this paper are:
\begin{itemize}\setlength\itemsep{0em}
\item We discuss a novel unifying perspective over CI, CB and UC problems, highlighting similarities and differences when considered as problems of learning with adaptively penalized losses. 
\item We introduce a new approach to tackle CI, CB and UC conditions, that relies solely on a modification of the training loss. Our approach applies generically to any classification problem (though for the sake of conciseness it is described for the binary case alone), and does not pose any restriction on the nature of the learning model. 
\item We provide an experimental assessment of the proposed method on both synthetic and real-world data, confronting our FBI-loss with state-of-the-art approaches from CI, CB and UC.
\item As a consequence of the previous point, this work shows an in-depth performance comparison of various methods to deal with different forms of unbalance, exploring many configurations of data type, task complexity and unbalance degree.
\end{itemize}

\section{Background}
In this section we briefly overview the research on CI, CB and UC problems. Synthesizing the contents of the following paragraphs, the strategies to deal with CI are mainly aimed at re-balancing the data or the error committed on them, while to address CB, the objective is focused on removing the dependency of the classification results or of the training data from the confounder; finally, to cope with UC, all approaches start from one definition of fairness and try to build a classifier that satisfies that definition.

\subsection{Class Imbalance}\label{CI_related}
CI can occur as the result of sampling issues or, possibly more often, for a research interest towards classifying rare events such as: defects \shortcite{rodriguez2014preliminary}, fraudulent behaviours \shortcite{olszewski2012probabilistic}, diseases \shortcite{yu2012mining} and natural calamities \shortcite{maldonado2014imbalanced}. Also Dense Object Detection (DOD) can be viewed as a CI problem \shortcite{lin2017focal}, further increasing the range of applications in which dealing with CI is fundamental.\\
According to a recent review, \shortcite{haixiang2017learning}, the approaches developed to cope with CI in the past decade can be broadly divided in: re-sampling techniques \shortcite{loyola2016study}, cost-sensitive learning \shortcite{krawczyk2014cost} and ensemble methods \shortcite{galar2013eusboost}.\\
Given the variety of applications affected by CI, re-sampling became a common choice for its ease of implementation, not requiring ML experts. A high number of intelligent methods adopting under-sampling, over-sampling and their combination have been developed to balance the classes while preserving valuable information for learning and avoiding over-fitting. However, an exhaustive comparison \shortcite{van2007experimental} among different sampling techniques showed that their effectiveness highly depends on both the learner and the problem domain. For instance, over-sampling is more difficult to apply to complex and structured data \shortcite{ando2017deep}, such as graphs or time-series.\\
Cost-sensitive learning, which addresses CI by weighing more the errors on the minority class, is computationally more efficient and flexible than re-sampling methods, but less popular \shortcite{haixiang2017learning}. The difficulty in setting the cost matrix values has been pointed out as a possible reason why cost-sensitive learning did not get the same attention \shortcite{krawczyk2014cost}. Nevertheless, common strategies to address this issue consist in setting the misclassification cost for the majority class to 1 while treating the one of the minority class as a hyperparameter, or by fixing it to the inverse class frequency ratio \shortcite{castro2013novel,lan2009joint}. This last option constitutes a mathematical re-balancing of the CI, as argued by \shortcite{shen2005loss,elkan2001foundations}.\\ 
Ensemble methods, being invented to broadly increase the generalizability of a classifier, are another widely adopted and effective solution to deal with CI. Bagging, boosting and hybrid ensemble learning have all been explored \shortcite{galar2013eusboost}, similarly to methods in which the various classifiers are trained using re-sampling techniques specifically designed to address CI \shortcite{sun2015novel,tian2011imbalanced}. Some works also consider to combine a cost-sensitive framework with ensemble learning \shortcite{fan1999adacost,ting2000comparative,sun2007cost}.\\
In DOD, instead, until a few year ago, the dominant paradigm was based on two-stage detectors \shortcite{uijlings2013selective,girshick2014rich,erhan2014scalable}, in which the first stage acts as a pixel filter, i.e. selecting a sparse set of pixels that should include all the objects to detect while discarding the majority of background, and the second one classifies them. While one-stage detectors had the potential to be faster and simpler, they trailed the accuracy of two-stage ones. In 2017, a turning point work \shortcite{lin2017focal} showed that DOD can be viewed as a CI problem between foreground and background, showing that the filtering stage, under-sampling the majority class, is what makes the two-stage detectors superior. This study, thus, suggested a cost-sensitive learning based approach to DOD, that down-weighs easy examples in order to focus training on hard-to-classify pixels. The effectiveness of this approach, which has been applied in a wide variety of applications \shortcite{sun2019drug,romdhane2020electrocardiogram,wang2018focal,pasupa2020convolutional,gao2020incremental,le2020transfer,zhang2018person}, has recently brought to the fore cost-sensitive learning for CI.

\subsection{Confounding Bias}\label{CB}
Observational studies had focused in removing confounding factors from data much before the ML community started turning its attention towards biases effects. Historically, this issue has been addressed by developing data-domain-specific normalization techniques or regressing out the influence of confounding variables from the data \shortcite{adeli2019bias}.\\
With the advent of ML and the possibility to collect large volumes of data, this concern fade out giving way to what is a posteriori known as the "big data hubris", i.e. 'the implicit assumption that big data are a substitute for, rather than a supplement to, traditional data collection and analysis' \shortcite{lazer2014parable}. A popular example of this phenomenon is given by the Google Flu Trends (GFT) \shortcite{GFTlink}, which was developed to predict the flu trend basing on searched terms on Google. Despite the initial enthusiasm, GFT showed periodical prediction fails caused by the large number of training data which increases the probability to find spurious correlations. For instance, it was found that a relevant search topic in one GFT version was high school basketball, because of its seasonal occurrence that was by chance associated to a previous flu peak \shortcite{ginsberg2009detecting}.\\
This and many other clamorous misleading examples \shortcite{caruana2015intelligible,brown2012adhd,zech2018confounding} brought the attention of the ML community to the importance of building classifiers that are robust to confounding variables to ensure generalization of the prediction performances.\\
To date, explainable algorithms \shortcite{ribeiro2016should,ross2017right}, causal models and graphs \shortcite{castro2020causality} can be used to understand whether the classifier learned a wrong or unexpected pattern. In addition, some specific tools to detect and quantify the confounding effect of a variable have been developed \shortcite{ferrari2020measuring,neto2018using}. To address and avoid learning from confounding biases, traditional normalization techniques, commonly employed in observational studies, turn out to be inefficient when their output is fed into ML algorithms. In fact, these algorithms can spot complex and distributed patterns, that are difficult to model and thus correct through normalization or regressions \shortcite{ferrari2020dealing}.
Instead, Invariant Feature Representation (IFR) constitutes a 'finer' and data-driven form of normalization, that is recently gaining attention to remove the dependence from confounders \shortcite{adeli2019representation,zhao2020training}. Other state-of-art techniques to address CB are based on Adversarial Learning (AL) \shortcite{adeli2019bias,kim2019learning}, which allows to maximize the ability of the classifier to predict the correct classes while constraining it to avoid learning to distinguish data on the basis of the confounder. Finally, approaches mixing IFR and AL have also been explored \shortcite{ganin2016domain,xie2017controllable}.   

\subsection{Unfair Classification}\label{UT}
Fairness issues in predictive modeling gained attention much before CB problems, for their societal impact.
A popular example of UC has been described by
Olver {\it et al.}~\shortcite{olver2014thirty}, who showed that the COMPAS algorithm, used by American courts to assess the offender's probability of recidivism, unfairly predicts higher risk scores for African-Americans.
This effect stems from the hidden biases in training data and labels, which are based on previous unfair judgments.\\
According to a recent review \shortcite{corbett2018measure} on fair machine learning, over the last years three broad definitions of fairness emerged: 
\begin{itemize}\setlength\itemsep{0em}
    \item Anti-classification (or unawareness): decisions are not based on protected attributes (e.g. race, gender and their proxies).
    \item Calibration (or individual-fairness): certain measures of predictive performance are independent on protected attributes conditional on a risk score, that is formulated based on prior knowledge of the task domain and that should guarantee similar outcomes to similar individuals.
    \item Classification parity (or group-fairness): certain measures of predictive performance (e.g. false positive and negative rates and others) are equal across the groups defined by protected attributes.
\end{itemize}
Based on these definitions, several pre-, in- and post-processing methods have been developed to enforce fairness in classification. \\
On anti-classification, simply removing protected attributes from training turned out to be inefficient, because of their indirect influence on the remaining ones \shortcite{pedreshi2008discrimination}. Thus, algorithms that prohibit the use of both protected attributes and their proxies have been explored \shortcite{grgic2016case,johnson2016impartial}.
However, besides the obvious difficulty in determining which features should be avoided, the main issue of anti-classification is that protected attributes or their proxies may even be essential to train a fair classifier. For instance, it is recognized that women, with respect to men with similar criminal history, have less probability to commit a future violent crime, thus a classifier that does not take gender into account would still be unfair \shortcite{corbett2018measure}.\\
The definition of calibration is instead less explored and it is usually implemented through post-processing methods which produce calibrated outputs from classification algorithms \shortcite{zhao2017men,hardt2016equality}.
The main limitation of calibration is that it simply shifts the evaluation of fairness from the comparison between outcome distribution and protected attributes to the one between outcomes and the distribution of a risk score, which can be defined with a certain degree of subjectivity. Thus, calibration alone does not ensure that decisions are equitable \shortcite{corbett2018measure}.\\
Finally, the definition of classification parity does not require any domain expertise such as the identification of proxies of the protected attributes or the definition of a risk score. For this reason, a high number of works focused on it by seeking to equalize the predictive performance across the groups defined by protected attributes through a constrained learning loss. In this context, the most popular fairness metrics used as constraints are:
\begin{itemize}\setlength\itemsep{0em}
\item Demographic Parity: it imposes an equal positive rate across groups. Despite being legally grounded, e.g. backed up by the legal ”four-fifth rule” \shortcite{bobko2004four}, demographic parity does not draw a distinction between false and true positives, a characteristic that makes it insufficient to guarantee fairness in several examples \shortcite{dwork2012fairness}.
\item Equalized Odds (EO): it guarantees both equal True Positive Rate (TPR) and False Negative Rate (FNR), which is equivalent in mathematical terms to the conjunction of equal False Positive Rate (FPR) and FNR \shortcite{verma2018fairness}.
\item Equal Opportunity: a relaxed version of EO that ensures only equal TPR, which is considered a sufficient condition for certain applications (e.g. hiring) \shortcite{hardt2016equality}.
\end{itemize}
Despite classification parity being the most popular fairness definition implemented in a ML context, it also suffers from statistical limitations \shortcite{corbett2018measure}. It can be concluded that approaches to address UC have historically focused on providing an algorithmic definition of fairness to be included in the classifier. This approach ends up becoming an unavoidable trade-off between enforcing one notion of fairness (sacrificing other desirable ones) and obtaining a better classification performance \shortcite{corbett2018measure}.

\section{Our approach: a unifying perspective and solution over CI, CB and UC.}
In this section, we describe in mathematical terms why CI, CB and UC can be viewed as different expressions of unbalance and how to address them with a unique solution inspired to cost-sensitive learning. First, we introduce our cost-based formalization of the problem, then we show how cost-sensitive learning, typically employed in CI problems, can be used as well to address the underlying unbalances characterizing CB and UC. Finally, we describe our FBI-loss.

\subsection{Weighted Cost Formulation}\label{sect:notation}
Let us define $(X, Y)$ the observed data in a two-class classification problem, where $X$ is the feature vector and $Y \in \{0,1\}$ is the target label, regarded as a realization of a binary random variable following a Bernoulli distribution. 
The unconditional probability of $Y=1$ is $\pi = P[Y=1]$, while the
conditional one given $X=x$ is $\eta(x)=P[Y = 1|x]$.\\
Let us suppose we want to attribute different costs, $c_0$ and $c_1$, to the errors made misclassifying the class $Y=0$ or $Y=1$ respectively. In this case, the expected cost of predicting the class 1 given an input $x$ is the product between the probability that the instance does not belong to the class 1 and the cost of misclassifying an element of the class 0: $c_0(1-\eta(x))$. As a consequence, the expected cost of predicting the class 0 is: $c_1 \eta(x)$.
Given these definitions, the optimal condition to predict the class 1 is when the cost of predicting 1 is lower than the one of predicting 0: 
\begin{align}
    c_0 \cdot (1-\eta(x)) &< c_1 \cdot \eta(x)\\
    \frac{c_0}{c_0+c_1} &< \eta(x) \label{eq2}\\ 
    c &< \eta(x) \label{eq3}
\end{align}
where we defined $c=c_0/(c_0+c_1)$. Clearly, $\eta(x)$ is unknown and has to be estimated from the data during the learning process, which consists in finding an estimated $p(x)$ (for conciseness $p$) that minimizes the cost-weighted misclassification error:
\begin{equation}\label{weighted-error}
    E(y,p) = (1-c)y\mathds{1}_{[p(x) \leq c]} + c(1-y)\mathds{1}_{[p(x) > c]} 
\end{equation}
where the function $\mathds{1}_{[input]}$ has value 1 when $input$ is true and 0 otherwise, and the conditions used as inputs are the rules for predicting respectively class 1 and 0, according to Eq. \ref{eq3}. However, given that $E(y,p)$ is not smooth, it is impossible to find its minimum through commonly employed gradient-based algorithms.
Thus, the search for $p(x)$ is usually performed using continuous approximations, such as the cost-weighted cross-entropy loss (H):
\begin{equation}\label{ce_cost}
    H\left(y,p\right) = - \left(1-c\right)y Log[p]-c\left(1-y\right)Log[1-p] .
\end{equation}
Note that, when $c_0=c_1$ meaning that $c=0.5$, $c$ becomes simply a constant factor multiplied to the second term of Eq. \ref{ce_cost} with no effects in the search for $p(x)$, thus the equation can simply be rewritten as the standard cross-entropy loss ($H^*$) with no weights:
\begin{equation}\label{ce}
    H^*\left(y,p\right) = -yLog[p]-(1-y)Log[1-p] .
\end{equation}
Until now, we have discussed the role of $c$ in weighing differently the errors on the two classes and, consequently, in the prediction process and in the definition of the learning loss.
However, $c$ can also be used to correct for the change of frequencies of the two classes, i.e. for correcting for CI at training. 
As mathematically derived by \shortcite{elkan2001foundations,shen2005loss}, a change in the value of the unconditional probability of $Y = 1$ from $\pi$ to $\hat{\pi}$, implies a change in the value of $c$, becoming $\hat{c}$, as described by:
\begin{equation}\label{baselinechange}
   \hat{c} = \frac{c\hat{\pi}(1-\pi)}{\hat{\pi}(c-\pi)-\pi(c-1)}.
\end{equation}
According to this equation, if we want to equally penalize errors on the two classes when they have the same unconditional probability (i.e. $c=0.5$ and $\pi=0.5$), but the composition of the training set is imbalanced, we should set $\hat{c}=\hat{\pi}$ to reach the same objective. For simplicity, this is the only situation we will consider throughout the paper: discussing the role of $c$ exclusively to correct for CI imposing $c=\pi$. To extend our dissertation to the case in which it is necessary to weigh the errors on the two classes differently also when $\pi=0.5$, it is sufficient to refer to Eq. \ref{baselinechange}.\\
For notational convenience, let us now simplify Eq. \ref{ce_cost} collecting the factor $c$ and introducing the constant $C=(1-c)/c$:
\begin{align}\label{ce_cost_2}
    H\left(y,p\right) 
    &= c\cdot \left( -\frac{1-c}{c} y Log[p] - \left(1-y\right)Log[1-p] \right) \\
    &\propto \bcancel{c} \cdot \Big( - C\: y\: Log[p] -  \left(1-y\right)\: Log[1-p] \Big) \label{eq9}\\
    &=  \Big( - C^y \: y\: Log[p] - C^y \: \left(1-y\right)\: Log[1-p] \Big) \label{eq10} \\
    &= C^y \cdot \Big(-y\: Log[p] -  \left(1-y\right)\: Log[1-p] \Big)\\
    &= C^y \cdot H^*\left(y,p\right) \label{weighted-ce&standard-ce}
\end{align}
Note that in Eq. \ref{eq9} $c$ can be removed because it does not depend on $p$ nor on $y$ and thus has no effect on the loss minimization process. In the passage between Eq. \ref{eq9} and \ref{eq10}, we can substitute $C^y$ to $C$ in the first addendum and introduce it in the second one, because the first addendum is 0 when $y=0$ and the second one is 0 when $y=1$.\\
Remembering that $C=(1-c)/c$ and that $c$ must be set equal to $\pi$ to correct for an imbalanced training set, we notice that $C=(1-\pi)/\pi=P[Y=0]/P[Y=1]$. This allows to give an easy interpretation to Eq. \ref{weighted-ce&standard-ce}: the cost-correction amounts to weigh the errors on class 1 with a factor corresponding to the inverse class frequency ratio in the training set; more elements of class 0 increase the weight of errors made on the class 1, and vice versa.

\subsection{Cost-sensitive learning: from CI to CB and UC}\label{BT}
In this section we want to show how the cost-sensitive learning framework, described to address CI problems, can be adapted to address other forms of unbalances: CB and UC. We build on the notation and formulation 
%Considering the notation and the binary classification problem already 
introduced in the previous section, but without class imbalance in the training set, i.e. $P[Y=1]=P[Y=0]$. Let us suppose to have access to another binary attribute $Z \in \{0,1\}$ (i.e., a possible confounder) of the observed data $(X,Y)$. Let us also suppose that the training examples are biased with respect to this variable, meaning that in the training set $P[Z = i|Y = i] > P[Z = j|Y = i]$, with $i\neq j$. In this biased configuration, if the classification task is too hard and distinguishing the data with respect to $Z$ is easier, the stronger is the inequality (i.e. the bias) the more the loss-minimizing $p(x)$ will be similar to $\zeta(x)=P[Z = 1|x]$. This misleading phenomenon, called either UC or CB depending on whether the attribute $Z$ is protected or not, happens because data with $z\neq y$ are under-represented in the training set; thus, errors made on them are overwhelmed by errors made on data with $z = y$.\\
To tackle this problem we should introduce in the loss function a relative error cost to penalize more those errors made on under-represented data, similarly to what previously described for CI problems.
To enhance the analogy between the CI and CB (or UC) problems we introduce a new attribute $U \in \{0,1\}$, given by the combination of the other two variables. In particular, we consider $u=|z-y|$, which is $1$ for  under-represented categories and $0$ for over-represented ones.\\
Thanks to this new variable, we can easily adapt Eq. (\ref{weighted-ce&standard-ce}) to define a new loss allowing to differentially weigh errors made on data with different $u$ values:
\begin{equation}\label{ce3}
    H'\left(y,z,p\right) = C^u\cdot H^* .
\end{equation}
In analogy with the previous discussion, $C$ should be set equal to $P[U=0]/P[U=1]$, i.e. the frequency ratio between the over-represented instances and the under-represented ones in the training set. 
%This approach can be easily extended to the case of multiple confounder, simply introducing new engegneered variables given by a combination of the task variable and the confounder that allows to define an under-represented and over-represented group of data. 

\subsection{The FBI-loss}\label{FBI_presentation}
\begin{figure}
\centering
\begin{subfigure}{0.45\textwidth}
 \centering
  \includegraphics[width=0.95\linewidth]{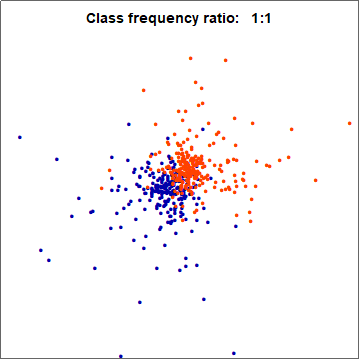}
  \caption{}
  \label{2Class}
\end{subfigure}
\begin{subfigure}{0.45\textwidth}
 \centering
  \includegraphics[width=0.95\linewidth]{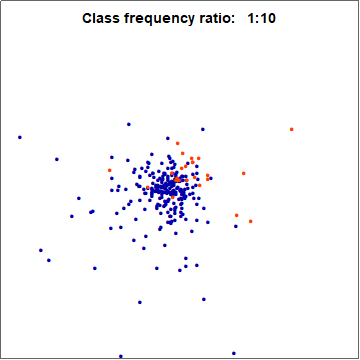}
  \caption{}
  \label{2Class_CI}
\end{subfigure}
\caption{Example of how CI can exacerbate task complexity. The blue and red dots represents examples of two hypothetical classes in their feature space. The two distributions are oppositely skewed but they are more densely populated in the overlapping region. When the imbalance is absent (Fig. \ref{2Class}) it is easy to recognize the two distributions, despite their significant overlap. When a strong imbalance is present (Fig. \ref{2Class_CI}), here obtained taking a random subset of the red class, it is more likely that the under-represented class contains exclusively points in the overlap region, thus, exacerbating the task complexity.}
\label{fig:2Class_CI_tot}
\end{figure}

In the last sections we have shown that CI, CB and UC can be viewed as different expressions of unbalance and that it is thus possible to extend our cost-sensitive learning framework to tackle also CB and UC. Now, we want to define a unique and effective loss to address all of them, by proposing an appropriate formulation for the $C$ factor of Eq. \ref{weighted-ce&standard-ce} and \ref{ce3}. \\
As shown by \shortcite{ferrari2020measuring} in CB problems, the misleading effect of a confounder variable ($Z$) is not only determined by the bias entity but also by the relative complexities of the actual classification task versus the task of recognizing the confounder. 
In fact, if distinguishing the two target classes ($Y$) is a harder task than discriminating with respect to $Z$, the classifier can easily learn a pattern based on $Z$ that allows to correctly predict the target classes of the data for which $Z$ and $Y$ are correlated (i.e., the over-represented data).\\
The role of task complexity is also important in CI problems, but in a subtler way. It can be observed that when the classes are well separated in the feature space the imbalance in training does not cause any misleading effect. At the same time when the classes are both sufficiently populated, imbalance is not a problem \shortcite{japkowicz2003class,joclass}. These phenomena suggest that the main issue behind CI is that the small dimension of the minority class exacerbate a problem of large overlap between the classes to distinguish, i.e. a hard classification task \shortcite{denil2010overlap} (see Fig. \ref{fig:2Class_CI_tot}).\\
Therefore we want to find a correction that takes into account the unbalance in the training set (being this a CI, CB, or UC case) and how it is amplified by the complexity of the task. This correction should help the loss to escape from wrong minima given by a badly conditioned training, but once it escapes, such a strong correction risks to overcorrect the unbalance. To avoid this effect we want also to introduce a modulating factor of the correction, that reduces the penalty when the prediction errors on the under-represented group become smaller. Given these considerations, we can formulate our FBI-Loss by modifying Eqs. \ref{weighted-ce&standard-ce} and \ref{ce3} as follows
\begin{equation}\label{fbiloss}
    FBI\left(y,d,p\right) = K^{d\cdot|y-p|\cdot\xi} \cdot H^*
\end{equation}
where the components of the loss are:
\begin{itemize}\setlength\itemsep{0em}
    \item $d$, which is the value of a dummy placeholder variable $D$ which stands for $U$, when the loss is used to address CB/UC, or for $Y$, when the loss is used for CI. In all cases, a value $d=1$ indicates that the data point belongs to the under-represented category. Being $d$ at the penalty exponent, the FBI-loss for over-represented data, which have $d=0$, reduces to $H^*$.
    \item $K$, which is determined by the dataset composition ($K=\frac{P[D=0]}{P[D=1]}$, as described in Sec. \ref{sect:notation}) and corrects exclusively for training unbalance.
    \item $|y-p|$, which modulates the penalty depending on the magnitude of the error made by the classifier. This factor can dynamically adjust the unbalance correction during training depending on the knowledge acquired. In other words, when the task is or has become particularly easy and the average errors made are small, $K^{|y-p|}$ tends to $1$, meaning that the bias correction is eliminated or dumped.
    \item $\xi$, a hyperparameter used to take task complexity into account. 
    \item $H^*$, the standard cross-entropy.
\end{itemize}
In principle, one could consider $K^{\xi}$ as a unique hyperparameter, however, we find that this distinction is useful to keep the bias and the complexity contributions separated, so that the knowledge acquired on $\xi$ can be transferred to similar problems.

\section{A Comparative Analysis of Related Works}
In this section we describe two methods specifically designed to address each one of the problems under examination (CI, CB and UC), for a total of six different approaches, with which we compare our FBI-loss both on real and synthetic data in our empirical analysis.

\subsection{Class Imbalance}
Being our plurivalent FBI-loss based on cost-sensitive learning, we compare its ability to address CI problems with other two cost-sensitive learning approaches developed for this kind of unbalance. In Section \ref{CI_related}, we state that one common implementation of cost-sensitive learning treats the weighing cost as a hyperparameter. We refer to this method as Cost Corrected loss (CC-loss):
\begin{equation}
   L_{cc}(y,p) = C^y \cdot H^*
\end{equation}
where $C$ is the hyperparameter and where we adopted the notation introduced in Section \ref{sect:notation} to enhance the differences with our FBI-loss.
Given that using $C$ as hyperparameter is equivalent to fixing $K$ and using its exponent $\xi$ as hyperparameter, the main difference with respect to our approach is in the presence of the modulating exponent $|y-p|$ which progressively reduces the effect of the penalization on the under-represented class, when the classifier is learning the correct pattern.\\
The second method we want to compare to is the Focal-loss \shortcite{lin2017focal} (F-loss), the already mentioned turning point work in DOD research, which can be written using our notation as follows:
\begin{equation}\label{focal}
   L_{F}(y,p) = K^y \cdot |y-p|^\alpha \cdot H^*
\end{equation}
where $K=\frac{P[Y=0]}{P[Y=1]}$ is a fixed factor with the same $K$ value used in the FBI-loss, and $\alpha$ is a hyperparameter. With respect to our formulation there are substantially two differences. First, in the F-loss, the role of $|y-p|$ is to decrease the loss magnitude on easy examples, regardless of whether they belong to the under-represented or over-represented group. Second, the effect of the hyperparameters is substantially different: $\xi$ linearly modifies $|y-p|$, which is applied exclusively to the under-represented group, while $\alpha$ is described as a 'focusing' exponent of $|y-p|$, non-linearly modifying the distinction between easy and hard examples.\\
We believe that for their similarity with our FBI-loss, the comparison with CC-loss and F-loss, can provide valuable information on which is the most effective way to decompose the CI problem. 

\subsection{Confounding Bias}
Following Section \ref{CB}, current state-of-the-art approaches to address CB are based on IFR and AL. Thus we compare our FBI-loss with two recent works using these methodologies. 
The first approach taken in consideration is an IFR work based on a Variational Autoencoder (IFR-VAE) \shortcite{moyer2018invariant} trained to maximize the following objective function:
\begin{equation}
    I(\Bar{x},y)-\beta I(x,\Bar{x})-\gamma I(\Bar{x},z)
\end{equation}\label{IFR}
where $x$ and $\Bar{x}$ are, respectively, the input features in the original space and in the new invariant representation, $y$ and $z$ are the class and confounding variables, and $I$ denotes mutual information. The terms $\beta$ and $\gamma$ are hyperparameters to be optimized. 
Maximizing this objective function means learning a feature representation $\Bar{x}$ in which the information related to the class is preserved (first term), while the one related to the confounder is destroyed (last term). The middle term imposes maximal compression of $x$ into $\Bar{x}$. The strength of this approach is that mutual information allows to detect, and thus remove, even complex dependencies from the confounder. However, given that the IFR-VAE is a pre-processing step, the optimization of the hyperparameters can be challenging because the effectiveness of the method can be evaluated only when the pre-processed data are used for a classification task.\\
The second approach with which we make a comparison is the Bias-Resilient Neural Network (BR-NN) \shortcite{adeli2019bias}, an in-processing solution that basically combines IFR with Adversarial Learning. The BR-NN is composed by a Feature Extractor (FE) and by two sub-networks fed with the outputs of FE. The first sub-network is trained to perform the classification task using the $H^*$ loss, while the second is trained to predict the confounding variable by maximizing the squared Pearson correlation $r^2$ between $z$ and its predictions.
The FE is trained in an adversarial way to minimize the quantity: $H^*-\delta r^2$.
Where $\delta$ is the only model hyperparameter.
This architecture allows to learn an invariant feature representation composed by data uncorrelated to the confounder but meaningful for the classification task, which should enable to achieve good performances avoiding confounding effects.\\
Both approaches are very different in nature, as well as in the code complexity, from our FBI-loss. They are based on the idea of 'learning' the confounder effect to remove it from the data (in the IFR case) or from the learning process (in the AL case). Our FBI-loss, instead, simply targets exiting the wrong minima given by the badly conditioned training configuration, without making nor imposing any assumption on what the classifier should learn.

\subsection{Unfair Classification}
As it emerges from Section \ref{UT}, EO represents a stable and widely applicable group-fairness definition which, in addition, has proved effective even if protected attributes are noisy \shortcite{awasthi2019effectiveness}.  
However, its implementation in a learning loss is not trivial and turns out to be non-differentiable \shortcite{cotter2018two}.
We want to compare the effectiveness of our FBI-loss with two widely different solutions that successfully incorporate an EO-based fairness definitions in the learner.\\
The first work we consider \shortcite{manisha2018neural} proposes to address the non-differentiability problem with a Proxy EO constraint, $C_{PEO}$, that uses the continuous predicted class probabilities instead of the categorical class predictions for calculating FPR and FNR. Their new loss (PEO-loss) consists in $H^*$ with an additive correction
\begin{equation}\label{eonn}
L_{PEO}= H^* + \lambda Max(0,( C_{PEO} - \epsilon  )),
\end{equation}
where $\lambda$ and $\epsilon$ are hyperparameters denoting the strength of and the tolerance on the EO correction, respectively.
Similarly to other approaches based on a fairness constraint, the learning process is basically a trade-off between accuracy (optimized by $H^*$) and fairness (imposed by the constraint), thus choosing an optimal value of the hyperparameter $\lambda$ is essential to avoid one term to overwhelm the other.\\
An elegant solution to simplify this search consists in solving the constrained optimization problem using Lagrangian multipliers, as proposed by the second work \shortcite{cotter2018two} with which we compare our approach. In their work, $\lambda$ is treated as one of the parameters to optimize during training. 
Basically, the Lagrangian optimization problem is formulated as a two player zero-sum game: the first player minimizes the objective function with respect to the model parameters $\theta$, while the second one maximizes it over $\lambda$ (or multiple $\lambda$s in the case of multiple constraints). This approach is well-founded because the loss differentiated with respect to the $\lambda$ coefficients does not contain the derivatives of the constraints. We will refer to this method as Lagrangian Fairness Optimization, or more simply \emph{LFO} method. Note that the pitfalls of this approach reside in the nesting of two alternate optimization phases, making the training process very sensitive to the learning rates.\\
Both methods chosen for comparison are based on EO constraints, but the search for the best compromise between classification performances and fairness is obtained with different ways. By comparing our approach with both of them, we can better understand the coarse advantages or disadvantages of a constraint-based approach with respect to a cost-sensitive one, regardless of the finer implementative details. 

\section{Experiments}
\begin{table}
\centering
\resizebox{\linewidth}{!}{
\begin{tabular}{|c|c|c|c|}
 \hline
 {\cellcolor{gray!20} \textbf{\textbf{Problem}}} &
 {\cellcolor{gray!20} \textbf{\textbf{Methods}}} &
 {\cellcolor{gray!20} \textbf{\textbf{Hyperparameter}}} &
 {\cellcolor{gray!20} \textbf{\textbf{Range}}} \\
 \hline
 \hline
 \multirow{2}{*}{Class Imbalance} & 
 CC & 
 C &
 1 - 1000 \\
 \cline{2-4}
 &
 F &
 $\alpha$ & 
 0 - 5\\
 \hline
 \hline
 
 \multirow{3}{*}{Bias} & 
 \multirow{2}{*}{IFR-VAE} & 
 $\beta$ &
 0 - 2 \\
 \cline{3-4}
 &
 &
 $\gamma$ & 
 0 - 2\\
 \cline{2-4}
 &
 BR-NN &
 $\delta$ &
 0 - 2 \\
 \hline
 \hline
 
 \multirow{5}{*}{Unfair classification} & 
 \multirow{2}{*}{PEO} & 
 $\epsilon$ &
 0 - 0.1 \\
 \cline{3-4}
 &
 &
 $\lambda$ & 
 0 - 2 \\
 \cline{2-4}&
  \multirow{3}{*}{LFO} & 
 $\epsilon$ &
 0 - 0.1 \\
 \cline{3-4}
 &
 &
 Learning rate 1 & 
 $10^{-6} - 10^{-3}$\\
  \cline{3-4}
 &
 &
 Learning rate 2 & 
 $10^{-6} - 10^{-3}$\\
 \hline
 \hline
 
 All & 
 FBI & 
 $\xi$&
 0 - 5 \\

 \hline

\end{tabular}}
 \caption{Competitive approaches analyzed in our comparative study. The last column shows the range of values explored for each hyperparameter in both the analysis on synthetic and the one on real-world data. Note that, despite all the methods have learning rates, the LFO is the only one being particularly sensitive to them, thus needing a fine tuning.}
 \label{table_hyp}
\end{table}
\subsubsection*{Objectives and experiments}
To evaluate the effectiveness and competitiveness of the proposed FBI-loss, we first test it on three synthetic datasets, each of them reproducing under controlled conditions one of the problems our loss was developed to address (i.e. CI, CB and UC).
For each of them, we compare the performances of our approach with the standard cross-entropy $H^*$ as baseline, and with two solutions from literature specifically developed for that problem. The second column in Table \ref{table_hyp} summarizes the methods under examination in our study.\\
This analysis on synthetic data allows to perform an exhaustive comparison, in which it is possible to study the performances of the various approaches with respect to different degrees of unbalance and task complexities, while avoiding problems that typically affect real-world datasets, such as scarce or poorly distributed data and their dependency from uncontrolled sources of variations.\\
To evaluate the applicability of our approach in practical scenarios, we replicate this comparison on three real-world datasets, each of them known in literature for being affected by one of the problems under examination.
In this second analysis, we can control the degree of unbalance by selectively choosing different subsets of the orginal data, but the complexity of the task is intrinsically determined by the nature of the data.\\
The code to replicate our empirical analysis is made publicly available \footnote{\url{https://bitbucket.org/elisaferrari/fbi-loss-code.git}}. 

\subsubsection*{Evaluation Metric}
In this paper we show how CI, CB and UC can be viewed as different expressions of unbalance in the training set. Thus, to easily evaluate the correction capabilities of our loss and of the other approaches under examination, we measure the performances of the models on two validation groups, respectively composed by data under-represented (UnderG) and over-represented (OverG) in the training set. In fact, if the model is misled by the unbalance we will see an increase of the performances on OverG with respect to UnderG that becomes more evident for larger magnitudes of unbalance.\\
For CI, UnderG and OverG are composed, respectively, by instances of the least and most numerous class. While for CB and UC problems UnderG is composed by instances with $u=1$ and OverG by the ones with $u=0$. Given that in the CI case UnderG and OverG contain elements of a single class, we quantitatively assess the validation performances using the Accuracy metric; in the other two cases we use the Area Under the Receiver Operating characteristic Curve (AUC).\\

We stress that, despite the adoption of Accuracy metric in CI problems is deprecated because its value is biased by the disproportion of the class frequencies \shortcite{branco2016survey}, we avoid this effect by computing the Accuracy separately for the UnderG and OverG groups. In fact, usually Accuracy is calculated as the ratio between the number of true predictions and the total predictions:
\begin{equation}
    Accuracy = \frac{true\;positives + true\;negatives}{all\;examples\;tested}
\end{equation}
equally weighing the positive and the negative instances even if they occur in different proportions in the test sample. In our evaluation, instead, we consider separately the quantities 
\begin{equation}
    Accuracy_{positive} = \frac{true\;positives}{all\;positives\;tested} \;\;\;\;\;
    Accuracy_{negative} = \frac{true\;negatives}{all\;negatives\;tested}
\end{equation}
Which provide unambiguous insights on the classifier performance.
\subsection{Synthetic data}

\subsubsection{Datasets description}
To simulate the CI problem, it is sufficient to build instances that depend on the target class $Y$ only, while for CB and UC they should also depend on the confounder variable $Z$.
Clearly, it is not possible to simulate the ethical implications that characterize the UC problem, thus the datasets for CB and UC are built in a similar way.\\
In our simulation, the synthetic instances are $N$-dimensional vectors, filled with a random white noise in the interval $[-\theta_{n},\theta_{n}]$.
Data dependency from a binary variable $V \in \{v_1,v_2\}$ is modeled using a variable-specific constant $\theta_V$ which is added to a set of $N_{v_i}$ features, that is different for the $v_1$ and $v_2$ values of $V$.
The sets of features used for the different values of $Y$ and $Z$ are all non-intersecting.\\
A careful measure of the complexity of the classification problem should take into account: $\theta_{n}$, $N$, the number of instances in training $N_T$, and all the $N_{v_i}$ and $\theta_V$ used.
In all our simulations we fix: $\theta_{n} = 5$, $N=100$, $N_T = 10^5$, $N_{v_i} = 4$ for all the $v_i$, and for CB/UC problems $\theta_Z = 3$. The only parameter we change to modulate the complexity is $\theta_Y$, which therefore becomes an inverse measure of complexity, i.e. the smaller $\theta_Y$ is, the harder the task is.

\subsubsection{Methods}
In this analysis, we compare the performances of our loss with related approaches from literature using synthetic datasets with varying degrees of unbalance and different task complexities.
Unbalance is computed as the ratio between the number of over-represented instances in the training set and its total size. As stated above, the inverse measure of task complexity is instead given by the parameter $\theta_Y$.\\
Besides $H^*$, that constitutes a baseline reference, all the other approaches to which we compare our loss (see Table \ref{table_hyp}) depend on one or more hyperparameters. To perform a fair comparison, when testing a model we let its hyperparameters vary for every combination of unbalances and complexities under consideration, and we select the configuration that maximizes simultaneously the performances on OverG and UnderG. The ranges of hyperparameters spanned are reported in the last column of Table \ref{table_hyp}.\\
Given the simplicity of the data, all the methods under study use a simple two-layer perceptron classifier comprising 50 and 10 hidden neurons.

\begin{figure}
\centering
\begin{subfigure}{0.45\textwidth}
 \centering
  \includegraphics[width=0.95\linewidth]{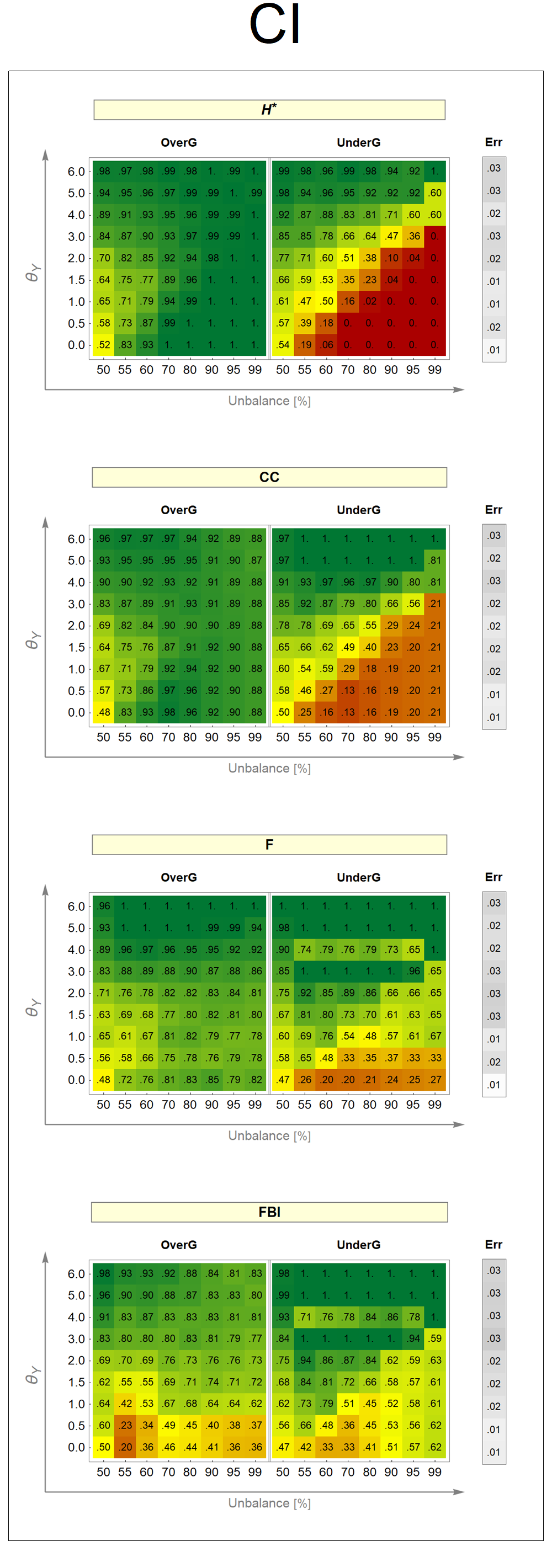}
  \caption{}
  \label{column_ci}
\end{subfigure}
\begin{subfigure}{0.45\textwidth}
 \centering
  \includegraphics[width=0.95\linewidth]{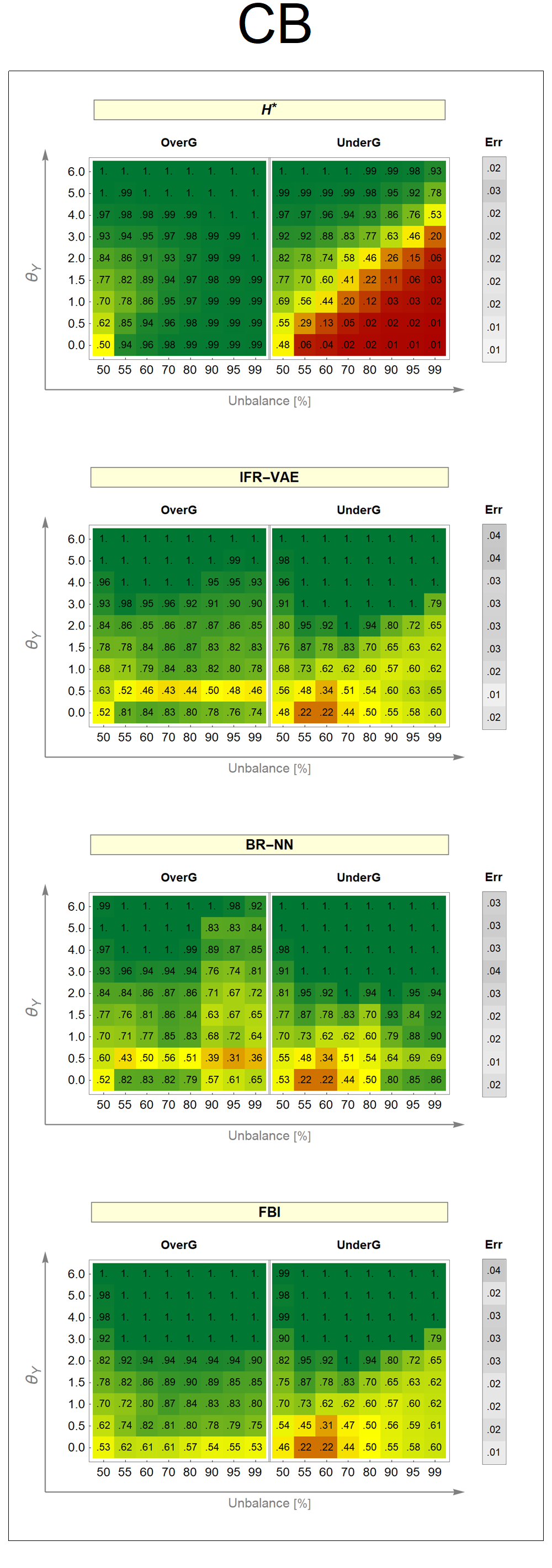}
  \caption{}
  \label{column_cb}
\end{subfigure}
\label{fig:columns_cicb}
\caption{Results of the analysis on synthetic data for CI problems (Fig. \ref{column_ci}). Results of the analysis on synthetic data for CB problems (Fig. \ref{column_cb}).}
\end{figure}

\begin{figure}
 \centering
  \includegraphics[width=0.45\linewidth]{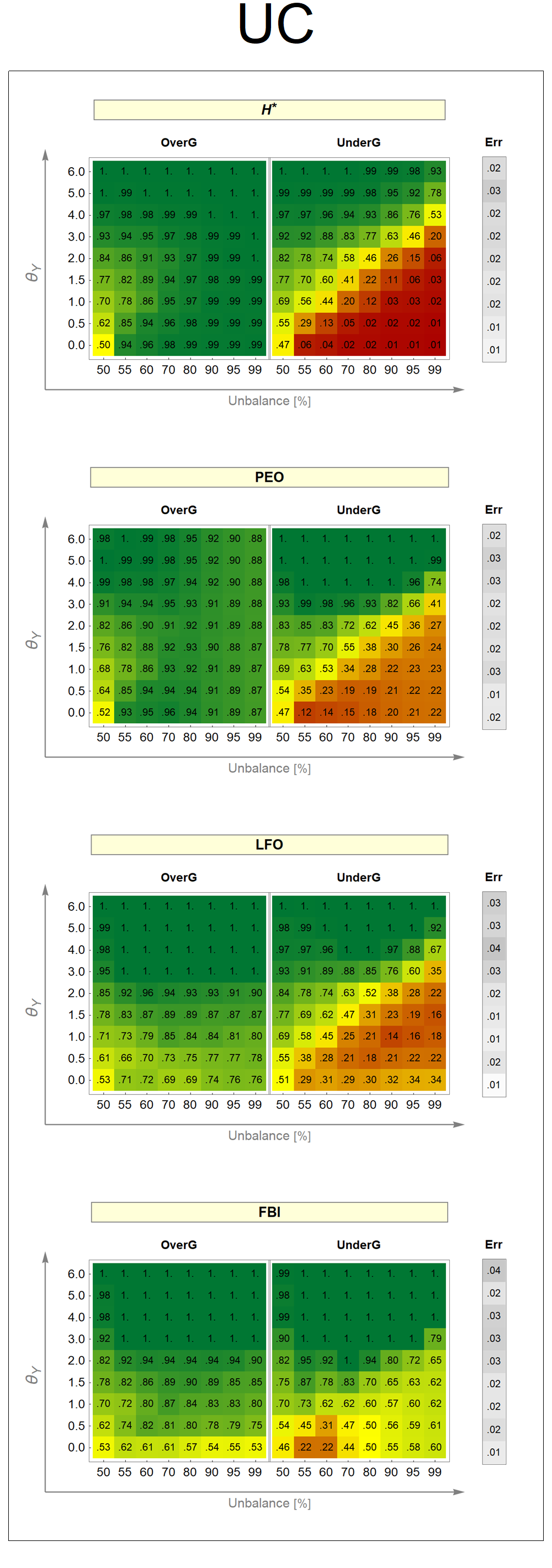}
  \caption{Results of the analysis on synthetic data for UC problems.}
  \label{column_uc}
\end{figure}

\begin{figure}
 \centering
  \includegraphics[width=1\linewidth]{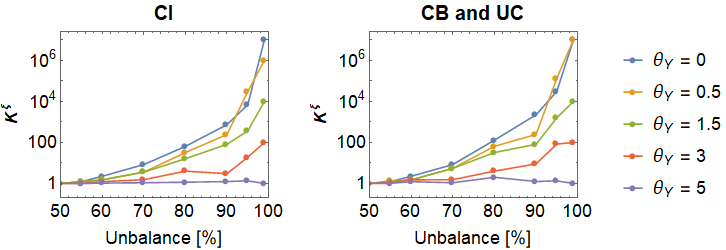}
  \caption{Best value for $K^{\xi}$ for different unbalances. Different task complexities are represented as different series in the plot.}
  \label{csi_plot}
\end{figure}

\subsubsection{Results}

\subsubsection*{Visualization of the results.}
To visualize the large amount of results produced with this analysis, we collect them in two separate matrices containing the mean AUCs (or Accuracies for CI simulations) computed on OverG and UnderG respectively (see Fig. \ref{column_ci}, \ref{column_cb} and \ref{column_uc}). Results are positioned inside the matrices in order of increasing unbalance along the x axis, and of increasing $\theta_Y$ (i.e. decreasing complexity) along the y axis. Each element of the matrix reports the mean performance obtained from 10 different training runs. Given the synthetic nature of the data and the high number of instances in validation, errors on the means are small. As a quantitative reference for this, we report the average standard deviation for the different task complexities in a grey column next to the matrices.\\
As already said, CB and UC synthetic datasets are obtained with the same procedure. Therefore, the results of $H^*$ and $FBI$ on these datasets are identical, but to facilitate the comparison between the approaches tested to solve the same problem, we report them twice.

\subsubsection*{Baseline reference.}
Before delving into the different results obtained in the CI, CB and UC simulations, let us consider the matrices of $H^*$ in Figures \ref{column_ci}, \ref{column_cb} and \ref{column_uc}. As it can be noted the behavior of the Cross-Entropy is the same for CI and CB/UC, which supports  the similarity between the two problems we discussed in this paper. In these figures, the column corresponding to unbalance = 50\% shows the ideal configuration in which performances depend exclusively on the complexity of the task, and they are roughly the same for OverG and UnderG. By increasing the degree of unbalance, the performances of $H^*$ diverge between OverG and UnderG, improving in the first case and drastically decreasing in the second one, until they become worse than a random guess. This divergence clearly starts for less severe levels of unbalance when the task is harder, which underlines the importance for a corrective loss to take into account the complexity of the problem.
The behaviour of $H^*$ gives us the key to interpret the remainder of the results: a good corrective method should produce two identical matrices with all columns being equal to the first one. \\

\subsubsection*{Class Imbalance}
Let us now analyze the CI problem (Fig. \ref{column_ci}). CC appears to operate better than a standard $H^*$ but the divergence between OverG and UnderG is still clearly visible. The F method produces better results but for higher complexities, e.g. $\theta_Y < 1.5$, it clearly shows an increase of performance in OverG and a specular decrease in UnderG. Note that for $\theta_Y = 4$, the UnderG matrix reports a light green line of difficult interpretation. Reanalyzing the results, we came at the conclusion that this line, such as other "strange" lines or spots, are caused by the fact that the range of hyperparameters explored did not allow to find the perfectly suited configuration for the specific combination of unbalance and complexity. This also explains why these anomalous results are usually clustered together.
Finally, FBI seems to be the most versatile approach. Despite a mild overcorrection visible in the easiest and hardest tasks, it is the one in which homogeneity is more preserved across columns and, remarkably, it is the only one avoiding to be drastically misled even for $\theta_Y = 0$, i.e., when the two classes cannot be distinguished.

\subsubsection*{Confounding Bias}
The interpretation of the results about CB problems (Fig. \ref{column_ci}) is straightforward. All methods under comparison (IFR-VAE, BR-NN, FBI) work well for most of the complexities and unbalances explored. The main difference across the three is for the impossible task: both IFR-VAE and BR-NN show mild opposite trends between OverG and UnderG, while FBI keeps its performances around the random guess, as expected.
The inferior performances of IFR-VAE and BR-NN on the impossible task may be related to the fact that both methods try to find a feature representation in which the dependence from classes is preserved and the one from confounders is destroyed. However, when $\theta_Y = 0$, data depend on class labels exclusively thanks to the confounder bias, thus this tug of war becomes more complex to handle.
In particular, in BR-NN the feature extractor is trained with a loss that penalizes the features for which the prediction of the class is highly correlated with the prediction of the confounder. This produces a mild but visible effect in the reported results. For high levels of unbalance in the OverG set, the confounder variable and the class label are highly correlated. Thus, also the features that depend exclusively on the class are penalized. This effect is made evident by the presence of a lighter rectangle on the right side of the OverG matrix.

\subsubsection*{Unfair classification}
On the UC problem, the plots in Fig. \ref{column_uc} do not require any further explanation. Both LFO and PEO are slightly better than $H^*$, but both are also clearly mislead by unbalance. FBI, consistently with the results on CI and CB, shows good correction capabilities.

\subsubsection*{The role of the hyperparameter $\xi$ in the FBI-loss}
Finally, to understand if $\xi$ effectively depends on task complexity, as originally hypothesized in the development of our FBI-loss, we report in Fig. \ref{csi_plot} the values of $K^\xi$ with $\xi$ optimized for each degree of unbalance and level of complexity for the CI and CB/UC problems. We can see that $K^\xi$ shows a clear increasing trend with task complexity,  bottom line being around $K^\xi=1$ for easier tasks. This means that the best configuration for the latter case is $\xi=0$, i.e. suppressing the $K$ correction.

\subsection{Real world data}

\subsubsection{Datasets description}
The three real-world datasets used in our analysis are: 
\begin{itemize}
    \item  Credit Card Fraud Detection \shortcite{dal2014learned} for CI problems. In this application the task is to distinguish between the imbalanced classes of fraudulent and non-fraudulent transactions.
    \item  ABIDE \shortcite{di2014autism} for CB problems. It is a multi-site Neuroimaging dataset, originally collected for the study of Autism. However, neuroimaging data were found to depend on many factors that can bias the learning process \shortcite{ferrari2020measuring}. In this study, we consider the task of recognizing gender from a 3D brain magnetic resonance image, when the data of the two classes are acquired using two different machines in different ratios.
    \item  UCI Adult Dataset \shortcite{Dua:2019} for UC problems. The task consists in predicting whether the income of a person exceeds $\$50K/yr$ or not and the UC problem arises from the gender bias distribution in the two classes.
\end{itemize}

\subsubsection{Methods}
With this second analysis we extend the comparison made on synthetic data to real-world applications. This means that the complexity is intrinsically given by the dataset under examination. Thus, to compare the various approaches we simply study the trend of the performance metric with respect to unbalance, making sure to use the same training and validation sets for all the methods tested on the same dataset. \\
Considering the different nature of the data and of the classification tasks in the three datasets, we use a neural network targeted to the nature of each dataset as described below:
\begin{itemize}
    \item CI: since the data used for this evaluation consist in small 1-dimensional feature vectors, for all methods under consideration we adopt a simple architecture comprising a single-layer perceptron with 16 hidden neurons.
    \item CB: in this application, data are 3D images; therefore, for the FBI loss we use a lighter version of AlexNet \shortcite{krizhevsky2012imagenet} (adapted to the 3D case) composed by 10 convolutional layers and 3 linear layers with a total of about 2 million trainable parameters. We analogously build the feature extractors of BR-NN and IFR-VAE with convolutional and linear layers.
    \item UC: Given the simplicity of both data and task, for all the approaches we use a single-layer neural network with 8 neurons. 
\end{itemize}

\subsubsection{Results}
The results of this analysis are more straightforward to visualise and interpret with respect to the ones described in the synthetic analysis.
A perfectly correcting model should show high and stable performances across the different unbalances on both UnderG and OverG.

\subsubsection*{Class Imbalance}
\begin{figure}
\centering
\includegraphics[width=1\linewidth]{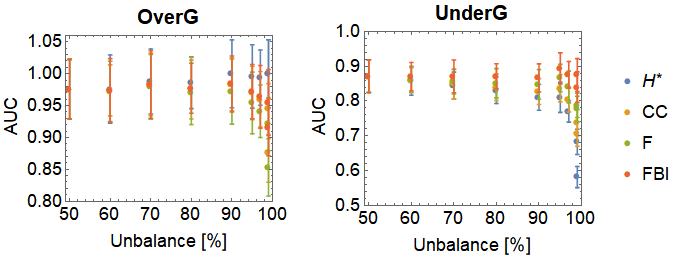}
\caption{Accuracy vs unbalance on OverG and UnderG for the CI problem.}
\label{fig:imbalance}
\end{figure}

The results for the CI analysis are reported in Fig. \ref{fig:imbalance}.
We can notice that the $H^*$ baseline shows the characteristic behaviour already described on synthetic data.
Given the relatively low complexity of the task, the performances of the various models are differentiated only for high values of unbalance. All the corrective methods show an improvement with respect to $H^*$ on UnderG, but suffer from a decrease in performance on OverG, probably due to over-correction. However, our FBI-Loss is the one with the most stable performances on both datasets.

\subsubsection*{Confounding Bias}
\begin{figure}
\centering
\includegraphics[width=1\linewidth]{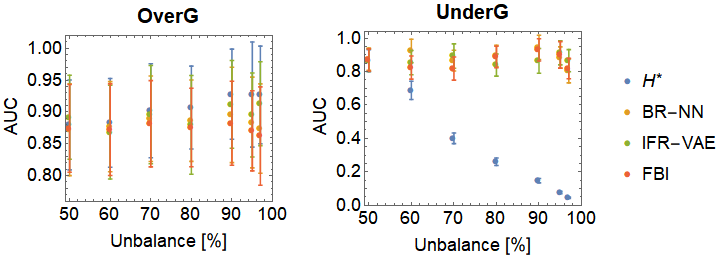}
\caption{AUC vs unbalance on OverG and UnderG for the CB problem.}
\label{fig:bias}
\end{figure}

Fig. \ref{fig:bias} shows the results obtained by the methods under CB. Consistently to what was observed in the synthetic analysis, all methods turn out to be perfectly correcting the CB problem, showing clear improvements with respect to $H^*$ and very stable trends. It is however remarkable that a loss-based method such as the FBI-Loss can be competitive with these much more complex and computationally impacting approaches on a real-world application. In fact, the FBI-loss is computationally efficient in two different ways: it has a single hyperparameter to be adjusted (requiring a significantly shorter exploration analysis at model selection), and does not require training of a separate classifier or encoder like those methods that adopt adversarial learning or variational autoencoders. In the experiments reported in this paper, the tuning and training process of the classifier using the FBI loss was between 2x and 10x more efficient than the IFR-VAE, BR-NN and LFO methods.

\subsubsection*{Unfair Classification}
\begin{figure}
\centering
\includegraphics[width=1\linewidth]{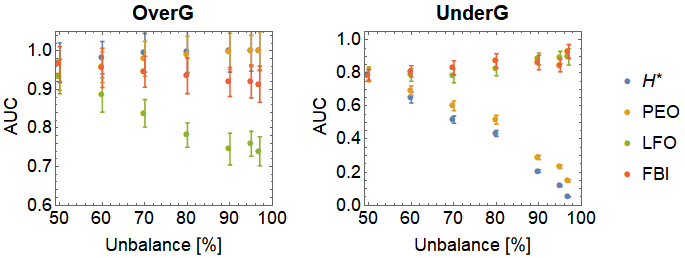}
\caption{AUC vs unbalance on OverG and UnderG for the UC problem.}
\label{fig:fairness}
\end{figure}

\begin{figure}
\centering
\includegraphics[width=1\linewidth]{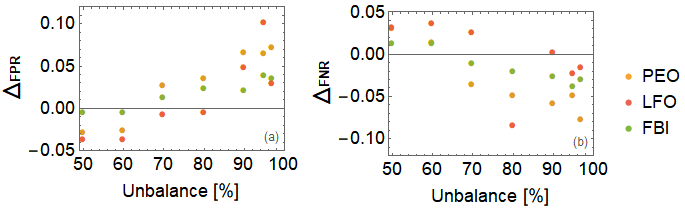}
\caption{FPR and FNR differences between the protected and unprotected classes with respect to unbalance, in the UCI dataset.}
\label{fig:fpr_fnr}
\end{figure}

The UC results are reported in Fig. \ref{fig:fairness}. We can notice that the PEO-Loss has significantly inferior correction capabilities, showing a behaviour that is very similar to that of $H^*$. LFO and FBI-Loss, instead, both present good performances, with the FBI-Loss being stabler on OverG.
Considering that the FBI-Loss has been developed as a general purpose cost-corrective function, we wanted to also verify whether it can indirectly ensure fairness. For this reason, for all the compared methods, we computed the FPR and FNR differences between the protected and unprotected groups on a validation dataset composed by the union of UnderG and OverG. The results are shown in Fig. \ref{fig:fpr_fnr}. As it can be noted, despite the FBI-loss was not specifically designed to satisfy EO constraint, and thus was not intended to minimize FPR and FNR, it achieves a similar degree of fairness with respect to the other methods, specifically designed to optimize these figures of merit. 

\section{Conclusions}
Class Imbalance, Confounding Bias and Fairness issues can severely undermine the usefulness of Machine Learning applications. In this work, we describe them from a unified perspective and we provide a unified formalization highlighting that they all arise from a different type of unbalance.
We reviewed the most recent solutions developed for each problem and suggested instead a unified cost-corrected loss (called FBI-loss) to address all these well-known learning problems at once. 
The main idea inherited by the Focal loss, specifically developed for Class imbalance, is to counteract unbalance by taking into account also the complexity of the classification problem.
Our FBI-loss is straightforward to implement, requires tuning of a single hyperparameter, and it is model-agnostic and general enough to be used with any classifier and for any kind of data. Furthermore, its only hyperparameter directly depends on the complexity of the task, which makes the tuning process easier for a data-domain expert.
When tested on both synthetic and real-world benchmarks, the FBI-loss shows performances that are competitive with those from relevant related approaches from literature and specifically tailored to tackle one single issue. Despite this preliminary formulation is suitable only for binary classification applications, the performance and versatility of the proposed loss should encourage future works in the direction of unified solutions.

%\input{Declaration.tex}

%\begin{acknowledgements}
%\end{acknowledgements}


\bibliographystyle{apacite}

%\bibliographystyle{abbrvnat}
%\bibliographystyle{spmpsci}
%\bibliographystyle{apalike}
\bibliography{Bibliography.bib}

\end{document}